\documentclass[10pt,journal,compsoc]{IEEEtran}
\ifCLASSOPTIONcompsoc
  \usepackage[nocompress]{cite}
\else
  \usepackage{cite}
\fi
\ifCLASSINFOpdf
\else
\fi

\usepackage{amsmath,amsfonts}
\usepackage{algorithmic}
\usepackage{algorithm}
\usepackage{array}
\usepackage[caption=false,font=footnotesize,labelfont=rm,textfont=rm]{subfig}
\usepackage{textcomp}
\usepackage{stfloats}
\usepackage{url}
\usepackage{verbatim}
\usepackage{graphicx}
\usepackage{cite}

\usepackage{times}
\usepackage{epsfig}
\usepackage{amssymb}
\usepackage{booktabs}
\usepackage{makecell}
\usepackage{multirow}
\usepackage{tabu}
\usepackage{diagbox}
\usepackage{bbding}
\usepackage{xcolor}
\usepackage{multicol}
\usepackage{color}
\usepackage{bm}
\usepackage{mathrsfs}
\usepackage{pifont}
\usepackage{wasysym}
\usepackage{psfrag}
\usepackage{indentfirst} 
\usepackage{comment}
\usepackage{enumerate}
\usepackage{threeparttable}
\usepackage{colortbl}

\begin{document}
%
\title{Heterogeneous Embodied Multi-Agent Collaboration}

\author{Xinzhu~Liu, Di Guo, and~Huaping~Liu
	\IEEEcompsocitemizethanks{\IEEEcompsocthanksitem Xinzhu Liu and Huaping Liu are with the Department of Computer Science and Technology, Tsinghua University, Beijing 100084, China, and also with the State Key Laboratory of Intelligent Technology and Systems, Beijing National Research Center for Information Science and Technology, Tsinghua University, Beijing 100084, China.
	\IEEEcompsocthanksitem Di Guo is with the School of Artificial Intelligence, Beijing University of Posts and Telecommunications, Beijing, China.
	\IEEEcompsocthanksitem Corresponding author: Huaping Liu (hpliu@tsinghua.edu.cn).}
}

\IEEEtitleabstractindextext{%
\begin{abstract}
Multi-agent embodied tasks have recently been studied in complex indoor visual environments. Collaboration among multiple agents can improve work efficiency and has significant practical value. However, most of the existing research focuses on homogeneous multi-agent tasks. Compared with homogeneous agents, heterogeneous agents can leverage their different capabilities to allocate corresponding sub-tasks and cooperate to complete complex tasks. Heterogeneous multi-agent tasks are common in real-world scenarios, and the collaboration strategy among heterogeneous agents is a challenging and important problem to be solved. To study collaboration among heterogeneous agents, we propose the heterogeneous multi-agent tidying-up task, in which multiple heterogeneous agents with different capabilities collaborate with each other to detect misplaced objects and place them in reasonable locations. This is a demanding task since it requires agents to make the best use of their different capabilities to conduct reasonable task planning and complete the whole task. To solve this task, we build a heterogeneous multi-agent tidying-up benchmark dataset in a large number of houses with multiple rooms based on ProcTHOR-10K. We propose the hierarchical decision model based on misplaced object detection, reasonable receptacle prediction, as well as the handshake-based group communication mechanism. Extensive experiments are conducted to demonstrate the effectiveness of the proposed model. The project's website and videos of experiments can be found at \textit{\url{https://hetercol.github.io/}}.
\end{abstract}

}

\maketitle

\IEEEdisplaynontitleabstractindextext

%
\IEEEpeerreviewmaketitle

\IEEEraisesectionheading{\section{Introduction}\label{sec:introduction}}

%
%
%
%


\IEEEPARstart{T}{he} embodied tasks are a series of comprehensive tasks integrating the process of environment perception, scene understanding as well as action execution, which are of great significance in the field of intelligent robotics. Recently, a number of embodied tasks have been studied to learn the relationship between visual perception and actions including visual navigation \cite{wortsman2019learning, yang2018visual, anderson2018vision}, instruction following \cite{shridhar2020alfred}, room rearrangement \cite{RoomR} and so on. However, these tasks are mainly studied in the single-agent scenarios, the work efficiency is likely to be low and the system is of poor fault tolerance. To solve the weakness of the single-agent setting in executing tasks, there are already some works investigating how to collaborate between multiple agents. Some simple disembodied environments such as the grid-world or 2D game scene are firstly used to learn the cooperation strategy \cite{omidshafiei2017deep, gupta2017cooperative, iqbal2019actor, zhou2020learning}. Recently, studies on multi-agent collaboration in the embodied visual environment have been conducted. \textit{FurnLift} \cite{jain2019two} and \textit{FurnMove} \cite{jain2020cordial} tasks are proposed to learn the tight collaboration in the action level between two agents. Then multi-agent exploration \cite{yu2022learning}, multi-agent visual navigation \cite{liu2022multi, wang2021collaborative}, multi-agent EQA \cite{tan2020multi} and multi-agent embodied task planning \cite{liuembodied} have been developed to study the cooperation strategy between multiple agents. But most of the multi-agent embodied tasks only consider the homogeneous agents, that is, all agents have the same ability. In practical tasks, different agents are likely to have different capabilities, and heterogeneous agents could leverage their different capabilities to deal with some complex problems that homogeneous agents cannot. The heterogeneous multi-agent task is common in real-world scenarios and has important practical applications. While the current research on heterogeneous agents is still limited to relatively simple experimental setups \cite{wakilpoor2020heterogeneous, mondal2022approximation}. In this paper, we concentrate on the collaboration strategy of heterogeneous agents in the embodied visual task.

\begin{figure}
	\centering
	\includegraphics[width=3.35in]{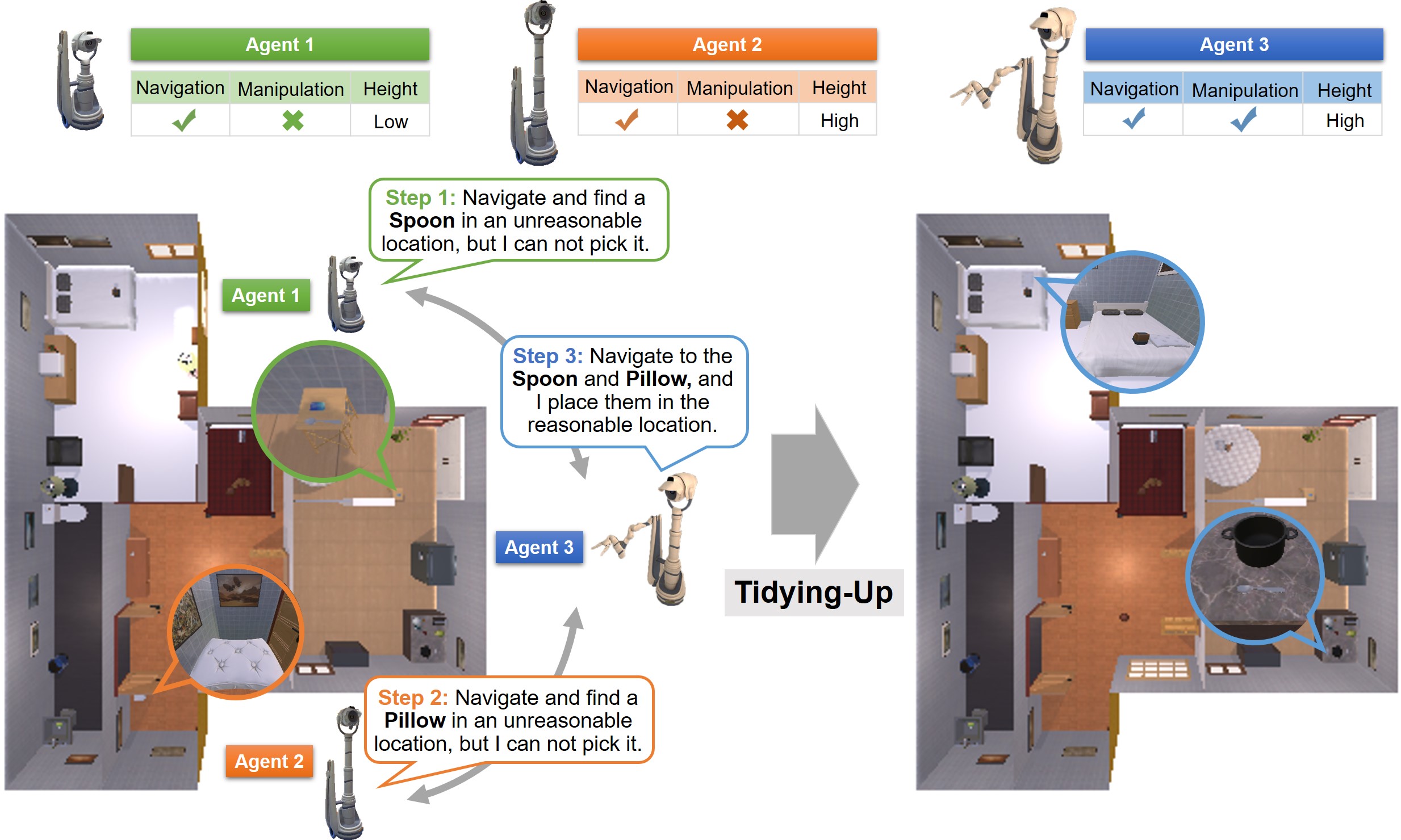}
	\caption{The overview of the heterogeneous multi-agent tidying-up task. \textit{Agent 1} and \textit{Agent 2} can only navigate, while \textit{Agent 3} is capable of both navigation and manipulation. The height of \textit{Agent 1} is low, while the heights of \textit{Agent 2} and \textit{Agent 3} are high. During exploration, \textit{Agent 1} and \textit{Agent 2} find a misplaced \textit{Spoon} and a misplaced \textit{Pillow} respectively but cannot pick them up, so they communicate with \textit{Agent 3}. Then \textit{Agent 3} navigates to pick up the \textit{Spoon} and \textit{Pillow}, and put them in reasonable places.}
	\label{fig:intro}
\end{figure}

To further study heterogeneous embodied multi-agent collaboration, verify and evaluate the proposed collaboration strategy, we propose the heterogeneous multi-agent tidying-up task, in which multiple agents need to find all misplaced objects and place them in reasonable locations. An overview of the proposed task is demonstrated in Fig. \ref{fig:intro}, in which multiple heterogeneous agents with different capabilities work together to detect the misplaced objects in the house and put them back to reasonable locations. The proposed task is quite challenging as it requires heterogeneous agents to make the best use of their own advantages and the complementary abilities of manipulation and navigation to conduct reasonable task planning as well as collaborate with each other to achieve the tidying-up task. Besides, this task also requires a deeper understanding of the scene. Agents need to analyze whether each object and its surroundings are placed reasonably and infer the reasonable receptacle to put the misplaced object based on visual observations and common sense reasoning without extra guidance. The most relevant work with ours is \textit{CH-MARL} in \cite{sharma2022ch}, but its experimental setup is relatively simple, since it only considers the collaboration among two agents, the humanoid and the drone, and agents can clearly know the target objects and receptacles without reasoning. The task has been completed in only 7 scenes. Our task studies collaboration among a larger number of agents moving on the ground in a larger number of indoor houses, and agents need to infer target receptacles themselves.

\renewcommand{\arraystretch}{1.0}
\begin{table*}[!t]
	\setlength{\tabcolsep}{1.1mm}
	\caption{Comparison Between Our Work and Other Related Embodied Tasks}
	\small
	\label{tab:comparison}
	\centering
	\begin{tabular}{ccccccc}
		\toprule[1pt]
		Task  & Multi-Agent & Heterogeneous Agents  &  Multi-Room Scene  & Navigation & Manipulation & Reasoning\\
		
		\Xhline{0.5pt}
		
		MQA \cite{deng2020mqa} & \ding{53} & \ding{53} & \ding{53} & \ding{53} & \checkmark & \ding{53} \\
		Rearrangement \cite{RoomR} & \ding{53} & \ding{53} & \ding{53} & \checkmark & \checkmark & \checkmark\\
		TIDEE \cite{sarch2022tidee} & \ding{53} & \ding{53} & \ding{53} & \checkmark & \checkmark & \checkmark\\
		Housekeep \cite{kant2022housekeep} & \ding{53} & \ding{53} & \checkmark & \checkmark & \checkmark & \checkmark\\
		MA-EQA \cite{tan2020multi} & \checkmark & \ding{53} & \ding{53} & \checkmark & \ding{53} & \ding{53}\\
		FurnLift \cite{jain2019two} \& FurnMove \cite{jain2020cordial} & \checkmark & \ding{53} & \ding{53} & \checkmark & \checkmark & \ding{53}\\
		CollaVN \cite{wang2021collaborative} & \checkmark & \ding{53} & \checkmark & \checkmark & \ding{53} & \ding{53}\\
		MA-Exploration \cite{yu2022learning} & \checkmark & \ding{53} & \checkmark & \checkmark & \ding{53} & \ding{53}\\
		MA-VSN \cite{liu2022multi} & \checkmark & \ding{53} & \ding{53} & \checkmark & \ding{53}  & \ding{53}\\
		MA-TaskPlannig \cite{liuembodied} & \checkmark & \ding{53} & \checkmark  & \checkmark & \checkmark & \checkmark\\
		$^*$CH-MARL \cite{sharma2022ch} & \checkmark & \checkmark & \checkmark  & \checkmark & \checkmark & \ding{53} \\
		Ours & \checkmark  & \checkmark  & \checkmark  & \checkmark & \checkmark  & \checkmark \\
		\toprule[1pt]
	\end{tabular}
	
	\begin{tablenotes}
		\item \footnotesize $^*$CH-MARL studies collaboration between two agents, the humanoid and drone, in indoor scenes, and agents can clearly know the target object and target receptacle without reasoning. Our work considers collaboration among multiple agents moving on the ground with or without the manipulator, and agents need to reason the reasonable target receptacle. 
	\end{tablenotes}
\end{table*}

In this paper, We propose a heterogeneous embodied multi-agent collaboration framework based on the hierarchical decision strategy and the handshake-based group communication mechanism. We evaluate the effectiveness of our model in the proposed heterogeneous multi-agent tidying-up task. We also build a benchmark dataset to complete the evaluation experiments in the heterogeneous multi-agent setting in multi-room scenes based on ProcTHOR-10K \cite{procthor}. The comparisons of our benchmark with other relevant embodied tasks are illustrated in Table \ref{tab:comparison}. The main contributions are summarized as follows:

\begin{itemize} 
	
	\item We propose a heterogeneous embodied multi-agent collaboration framework, which makes full use of the different capabilities of heterogeneous agents to collaboratively complete the whole task.
	
	\item We develop a handshake-based group communication mechanism for heterogeneous multi-agent collaboration. A hierarchical decision strategy based on scene understanding, scene reasoning with prior knowledge, and communication mechanism is proposed to solve the heterogeneous multi-agent collaboration task.
	
	\item We propose the heterogeneous multi-agent tidying-up task to evaluate the proposed heterogeneous collaboration strategy, in which multiple heterogeneous agents need to find all misplaced objects and place them to reasonable locations. We build a novel benchmark dataset to evaluate our model, which includes \textit{Single-Room} and \textit{Cross-Room} settings. We also provide the method to generate the task data in indoor houses for the tidying-up task. Experimental results demonstrate the effectiveness of the proposed heterogeneous multi-agent collaboration framework.
	
\end{itemize}

This paper is organized as follows. In Section \ref{sec:related}, the related existing researches are summarized. In Section \ref{sec:formulation}, the heterogeneous multi-agent collaboration framework and heterogeneous multi-agent tidying-up task are formulated. In Section \ref{sec:dataset}, we demonstrate the newly built dataset. The proposed heterogeneous multi-agent model is described in Section \ref{sec:method}. Then the experiments are shown in Section \ref{sec:experiment}. Finally, we come to the conclusion in Section \ref{sec:conclusion}.

\section{Related Work}\label{sec:related}

\subsection{Embodied Tasks}

In recent years, embodied visual tasks have attracted a large number of researches. Researchers have proposed and studied the visual navigation task including visual semantic navigation \cite{wortsman2019learning, yang2018visual, pal2021learning} and visual-and-language navigation \cite{anderson2018vision, fried2018speaker, wang2019reinforced, Wang2021vision, Lin2022adversarial, Qiao2023hop}, visual exploration \cite{chaplot2020learning}, remote embodied visual referring expression \cite{qi2020reverie}, embodied question answering (EQA) \cite{das2018embodied, gordon2018iqa, wijmans2019embodied, Luo2023depth, Tan2023knowledge} and instruction following \cite{shridhar2020alfred, zhang2021hierarchical}. Recently proposed visual room rearrangement \cite{RoomR} and tidying-up tasks \cite{sarch2022tidee, kant2022housekeep} require the agent to learn the abilities of deep scene understanding and scene reasoning in long-horizon tasks. However, the above tasks are mainly studied in single-agent scenarios, which may lead to low efficiency and poor fault tolerance. Researchers have further studied multi-agent embodied tasks, including collaborative object movement \cite{jain2019two, jain2020cordial}, collaborative exploration \cite{yu2022learning}, multi-agent visual navigation \cite{liu2022multi, wang2021collaborative}, collaborative EQA \cite{tan2020multi} and collaborative task planning \cite{liuembodied}. All these works consider homogeneous agents with exactly the same capabilities.

\subsection{Heterogeneous Multi-Agent Tasks}

Different from homogeneous agents, heterogeneous agents can leverage their different capabilities to allocate corresponding sub-tasks and cooperate to complete complex tasks. Heterogeneous multi-agent tasks are common in real-world scenarios and have important practical applications, such as the air-ground cooperative system, collaborative manipulators with different functions in the factory, etc. Recently, researchers have proposed heterogeneous multi-agent reinforcement learning (MARL) methods to solve different tasks including known environment mapping \cite{wakilpoor2020heterogeneous}, cooperative cache in cognitive radio networks \cite{gao2022cooperative}, autonomous separation assurance \cite{brittain2022scalable}. Then inductive heterogeneous graph MARL \cite{yang2021ihg}, heterogeneous MARL with mean field control \cite{mondal2022approximation}, and meta RL-based approach \cite{chen2021meta} have been proposed to improve the method. In this paper, we consider the collaboration and communication among heterogeneous agents in embodied tasks, in which all heterogeneous agents simultaneously perceive the environment and collaboratively perform actions to complete the task. It is different from \textit{CoMON} \cite{patel2021interpretation}, in which one oracle agent only provides information without executing actions and only one agent executes the task. \textit{CH-MARL} considers the task of heterogeneous agents in the embodied environment with two different robots including humanoid and drone \cite{sharma2022ch}, but the experimental setup is relatively simple, in which agents only need to move an object to a specific receptacle without the requirement of scene reasoning. Besides, \textit{TarMac} \cite{das2019tarmac} and \textit{When2com} \cite{liu2020when2com} are also related to our work on multi-agent communication, but they only study the communication of homogeneous agents without considering different abilities of heterogeneous agents. \textit{TarMac} broadcasts the signature and value vectors to others, \textit{When2com} utilizes a handshake-based method to complete the intra-group communication, while our model develops the handshake-based group communication mechanism with both intra-group and inter-group communication, which is effective in collaboration of heterogeneous agents.

\section{Problem Formulation}\label{sec:formulation}

For the heterogeneous multi-agent collaboration tasks, we consider that there exist $N$ heterogeneous agents ${A}^{(1)}, {A}^{(2)},\cdots,{A}^{(N)}$. Each agent $A^{(i)}$ has a capability representation $B^{(i)}$, which consists of three different dimensions, namely the perception ability $P^{(i)}$, the action ability $E^{(i)}$ and the agent morphology $H^{(i)}$, $B^{(i)}=(P^{(i)},E^{(i)},H^{(i)})$. Specifically, in the heterogeneous multi-agent tidying-up task, each agent has the same visual perception ability. The action ability $E^{(i)}$ can be indicated as $E^{(i)}=(Nav^{(i)}, Mani^{(i)})$, where $Nav^{(i)}$, $Mani^{(i)}$ denote whether the agent has the ability of navigation and manipulation respectively. If the agent has the corresponding ability, its value is 1, otherwise, the value is 0. As for the agent morphology $H^{(i)}$, we consider different heights of the agents and represent it as $H^{(i)} = (Hei^{(i)})$, where $Hei^{(i)}=1$ denotes the height of the agent is high, roughly 0.9 meters, and $Hei^{(i)}=0$ denotes the height of the agent is low, roughly 0.22 meters. $N$ heterogeneous agents have different $(E^{(i)},H^{(i)})$, and different action spaces $I^{(i)}$. Specifically, if $Nav^{(i)}=1$ and $Mani^{(i)}=0$, $I^{(i)} = I_{nav}$. If $Nav^{(i)}=1$ and $Mani^{(i)}=1$, $I^{(i)} = I_{nav} \cup I_{mani}$, where $I_{nav}=\{MoveAhead,\; \allowbreak
MoveRight, \; \allowbreak
MoveLeft, \; \allowbreak
RotateRight,\; \allowbreak
RotateLeft,\; \allowbreak
LookUp,\; \allowbreak
LookDown,\;
Stop\}$ denotes the navigation action set and $I_{mani}=\{PickUp, \; \allowbreak
PutDown, \; \allowbreak
Drop\}$ represents the manipulation actions.

There exist $M$ objects in the house which are denoted as $\mathcal{O}=\{{o}_1, {o}_{2},\cdots,{o}_{M}\}$. Among these $M$ objects, the subset $\mathcal{O}_{pick} \subseteq \mathcal{O}$ denotes the set of objects which can be picked up by the agent. The subset $\mathcal{O}_{recep} \subseteq \mathcal{O}$ represents the set of objects which can be used as a receptacle and other objects can be placed on their surface or inside. The location of the $j$-th object can be represented as $(o_j, p_j, r_j)$, where $p_j \in \mathcal{O}_{recep}$ denotes the receptacle that $o_j$ is placed in, and $r_j$ is the room where $o_j$ and $p_j$ are in. The environment can provide a judgment of whether the current object is in a reasonable location with a discriminator $D$. If $D(o_j, p_j, r_j)=True$, $o_j$ is in a reasonable location, otherwise, $o_j$ is in an unreasonable location. There exist $k$ objects which are in unreasonable locations in the house, namely misplaced objects, and agents do not know the value of $k$. At the $t$-th step, the agent $A^{(i)}$ obtains the perception observation $e_t^{(i)}$ and the communication information $c_t^{(i)}$ from other agents. Each agent $A^{(i)}$ needs to learn a action strategy ${\pi}^{(i)}$ to generate the action $a_t^{(i)}$ from its action space $I^{(i)}$, $a_t^{(i)} = {\pi}^{(i)}(e_t^{(i)}, c_t^{(i)})$. The objective of this tidying-up task is to learn the optimal collaboration action strategy ${\pi}^{(i)}$ for each agent such that they can collaboratively find all misplaced objects and place them in reasonable locations to make $D(o_j, p_j^{*}, r_j^{*})=True, j = 1,2,\cdots,M$ with as few steps as possible, where $(o_j, p_j^{*}, r_j^{*})$ denotes the location of objects after all agents complete their actions and stop.

\begin{figure*}
	\centering
	\includegraphics[width=6.8in]{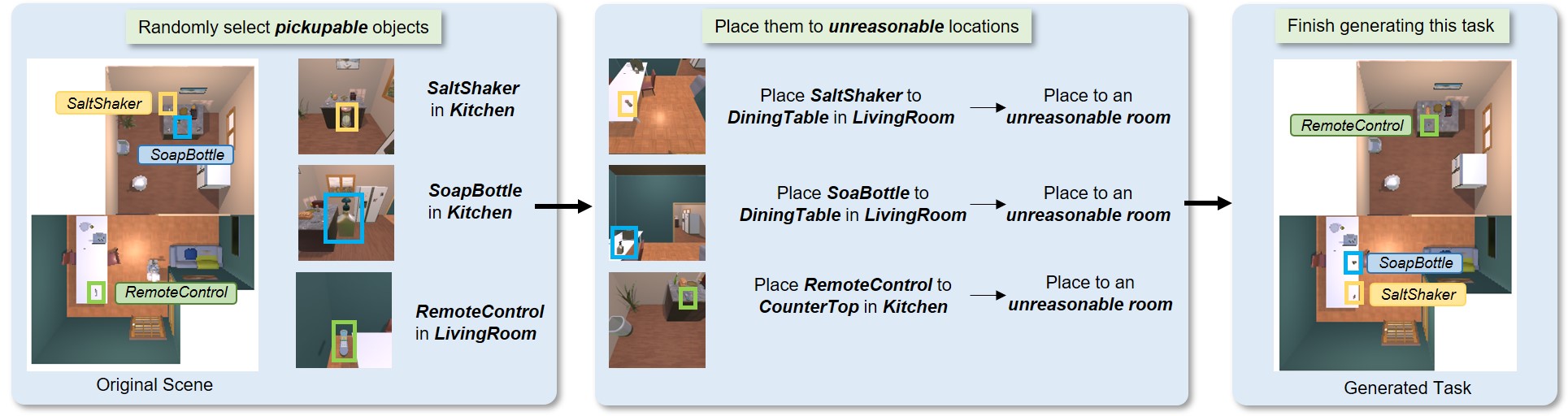}
	\caption{The process of generating the tidying-up task from the original scene. For each tidying-up task, we randomly select $k$ objects from the original house in ProcTHOR-10K and then change their current locations to unreasonable receptacles or unreasonable rooms to generate the task data.}
	\label{fig:process}
\end{figure*}

\section{Dataset}\label{sec:dataset}

To evaluate the proposed heterogeneous multi-agent collaboration model, we create the novel dataset for the heterogeneous multi-agent tidying-up tasks based on the ProcTHOR-10K \cite{procthor} which provides a large number of houses with multiple rooms. In each task, several objects are placed in unreasonable locations, and multiple heterogeneous agents are required to find these misplaced objects and put them to the reasonable locations. The newly built dataset contains the \textit{Single-Room} (\textit{Single}) and \textit{Cross-Room} (\textit{Cross}) tasks to train and evaluate the proposed model.


\begin{figure}
	\centering
	\subfloat[The proportion of the number of misplaced objects.]{\label{fig2a}
		\includegraphics[width=1.26in]{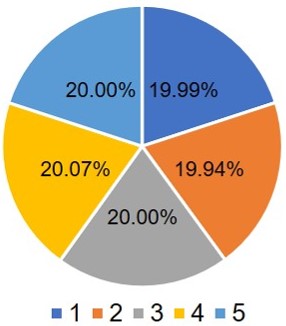}}
	\hspace{0.3in}
	\subfloat[The proportion of \textit{Single-Room} and 
	\textit{Cross-Room} tasks.]{\label{fig2b}
		\includegraphics[width=1.58in]{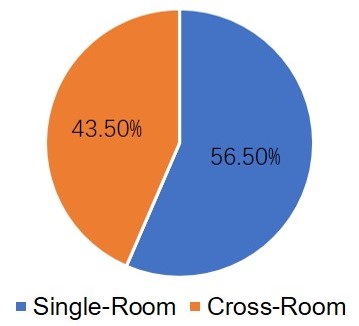}}
	\caption{The statistics of the built dataset.}
	\label{fig:statics}
\end{figure}


\begin{figure*}
	\centering
	\includegraphics[width=6.4in]{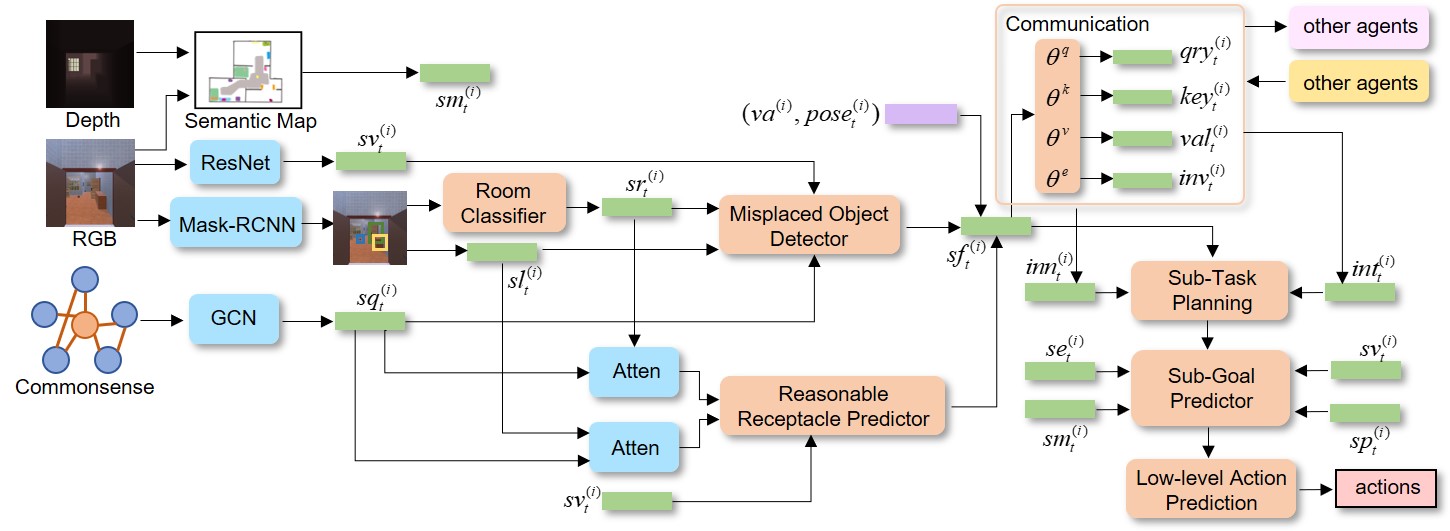}
	\caption{An overview of the structure of the proposed heterogeneous multi-agent collaboration model. The model consists of the misplaced object detector, the reasonable receptacle predictor, the communication module, and the hierarchical decision. The misplaced object detector judges whether there exists an object placed in an unreasonable location. The reasonable receptacle predictor generates a reasonable location to place the misplaced objects. The communication module shares communication messages among heterogeneous agents with the handshake-based group communication strategy. The hierarchical decision module predicts the next sub-task, sub-goal, and next low-level actions for each agent.}
	\label{fig:model}
\end{figure*}

ProcTHOR-10K is a generated dataset of houses with multiple rooms \cite{procthor} built based on AI2-THOR \cite{ai2thor}, including 10,000 training houses, 1000 validation houses, and 1000 testing houses, and each room belongs to one type of \textit{Kitchen}, \textit{LivingRoom}, \textit{Bedroom} and \textit{Bathroom}. We propose a method to generate multi-agent tidying-up tasks in ProcTHOR-10K. Since every single room in ProcTHOR-10K is built based on AI2-THOR, we regard the object placement relationships in AI2-THOR as the reasonable placement constraints in ProcTHOR-10K. We obtain the properties of objects and their reasonable locations from ProcTHOR-10K. Specifically, we obtain the objects which have the properties \textit{Pickupable} and \textit{Receptacle} to form the set $\mathcal{O}_{pick}$ and $\mathcal{O}_{recep}$ respectively. For each object type in $\mathcal{O}_{pick}$, we generate the triples of reasonable candidate locations $Re_i = \{(o_i, p_j, r_j)\}$ from the metadata, where $o_i \in \mathcal{O}_{pick}$, $p_j \in \mathcal{O}_{recep}$, $r_j \in \{Kitchen, LivingRoom, Bedroom, Bathroom\}$, $o_i$ can be put on $p_j$ in room $r_j$. We generate 10 meta-tasks for each house in ProcTHOR-10K, and the process is shown in Fig.\ref{fig:process}. For each tidying-up meta-task, we randomly select $k$ objects to change their current location $(o_i, p_i^{cur}, r_i^{cur})$ to an unreasonable location $(o_i, p_i^{new}, r_i^{new})$, and $k$ is randomly chosen from $\{1,2,3,4,5\}$. We use the action $PutObject$ in ProcTHOR-10K to place the misplaced object to an unreasonable receptacle or use the action $DropHandObject$, $ThrowObject$ to drop the object to a random location. For each misplaced object, $p_i^{new}$ and $r_i^{cur}$ may be the same or different. If $r_i^{new} = r_i^{cur}$, the misplaced object may be placed in an unreasonable location in the same room. If $r_i^{new} \neq r_i^{cur}$, the misplaced object may be placed to the location in another unreasonable room. 
All meta-tasks are divided into \textit{Single} and \textit{Cross} tasks. In \textit{Single} tasks, all $k$ misplaced objects are placed in the same room as their initial positions, and in \textit{Cross} tasks, at least one misplaced object is placed in a different room from its initial position. The statistics of the dataset are shown in Fig. \ref{fig:statics}. The proportion of the selected number of misplaced objects in the generated dataset is shown in Fig. \ref{fig2a}, and Fig. \ref{fig2b} illustrated the proportion of \textit{Single-Room} tasks and \textit{Cross-Room} tasks.

For each meta-task, we randomly generate 5 initial positions for multiple heterogeneous agents. We also generate the expert demonstration with heuristic shortest path methods for imitation learning. Specifically, we obtain the metadata and scene information of each task, and collect labels for the sub-task and sub-goal predictors of the action strategy module. We generate oracle sub-task lists for each agent from task information and find viewpoints that agents can interact with misplaced objects as target locations for manipulation. We generate the sub-task allocation results for each agent at each step according to the distance between each agent and misplaced objects. Then, we generate the heuristic shortest path in the sub-goal level, generate 5 reasonable trajectories per training task, and obtain pairs of state features and sub-goal actions ($\Delta x_t, \Delta y_t, \Delta rot_t, ope_t, stop_t$) from them, which are mentioned in Section \ref{sec:method}. Then we use sub-goal actions as labels for imitation learning.


We notice that our dataset is different from that in \cite{sarch2022tidee}. The tasks in our dataset are more difficult than \cite{sarch2022tidee}, and our tasks require deeper scene understanding and reasoning. Firstly, the tidying-up tasks in our dataset are built on large houses with multi-rooms and misplaced objects are likely to be placed in another unreasonable room, while the dataset in \cite{sarch2022tidee} are only built in the single room in AI2-THOR. Secondly, we consider the relationship among pickupable objects, receptacles and rooms, which means that the misplaced objects may be placed in an unreasonable receptacle, or in a reasonable receptacle but an unreasonable room. But in \cite{sarch2022tidee}, misplaced objects are generated only by being thrown away on the ground in the same room, and it does not need to consider the relationship between objects and rooms. It does not consider the relationship between objects and receptacles either. Besides, our dataset is used to solve the multi-agent tasks and each task has initial positions for multiple agents, while the dataset in \cite{sarch2022tidee} is designed for the single-agent task. Furthermore, the major differences between our dataset and \textit{Housekeep} in \cite{kant2022housekeep} are that our dataset has a larger scale, can be automatically generated through the proposed program, and is designed for heterogeneous multi-agent collaboration. We build a benchmark dataset based on ProcTHOR-10K which contains 10K houses with multiple rooms, while dataset in \cite{kant2022housekeep} only has 14 scenes. We also provide the program to generate task data and our dataset can be automatically generated through the program, while the generation of \textit{Housekeep} needs a large number of efforts from human annotations. Our program can also be easily extended to any other indoor simulation platform with a large number of houses to generate the corresponding task data. Meanwhile, our dataset is designed for heterogeneous multi-agent task, while \textit{Housekeep} is designed for the single-agent task. Misplaced objects in our dataset are likely to be placed in the receptacle or thrown on the floor, which are easier to be observed by agents with high or low height respectively, making it more suitable for the study of heterogeneous multi-agent collaboration.



\section{Methodology}\label{sec:method}

The proposed model that solves the multi-agent tidying-up task consists of four main modules: the misplaced object detector, the reasonable receptacle predictor, the communication module, and the hierarchical decision. The overview of this model is shown in Fig. \ref{fig:model}. The misplaced object detector judges whether there exists an object placed in an unreasonable location. The reasonable receptacle predictor generates a reasonable receptacle and room type to place the misplaced objects. The communication module transmits the communication information to other heterogeneous agents. The hierarchical decision module predicts the next sub-task, sub-goal, and next actions for each agent to execute.

\subsection{Misplaced Object Detector}

Each agent builds a top-down semantic map with the input RGB and depth using the method similar to the semantic mapping in \cite{liu2022multi}. At each sub-goal step, the agent obtains a local semantic map of size $G \times G \times (K_{total} + 2)$, where $G$ indicates the size of the local map, and $K_{total}$ is the number of object categories. Then agent $A^{(i)}$ generates the map embedding $sm_t^{(i)}$ with a pre-trained scene encoder similar to that in \cite{liuembodied} consisting of multiple convolutional layers.



We filter out spatial relationships among objects, receptacles and room types existing in our dataset from Visual Genome \cite{krishna2017visual} to form a graph structure. We regard these relationships as the commonsense prior knowledge. We use the Glove embedding \cite{pennington2014glove} to encode the class label, concatenate with the visual embedding as node features, and use GCN \cite{KipfW2017semi} to obtain the embedding of the commonsense $sq^{(i)}_t$. ResNet \cite{he2016deep} is utilized to extract the visual embedding $sv^{(i)}_t$ of the visual observation. Meanwhile, the pre-trained Mask-RCNN \cite{he2017mask} is used to detect the existing objects and obtain the feature $sl^{(i)}_t$. We construct a room classifier to predict the room type where the agent is currently located based on detected objects and their placement relationships. The classifier extracts the room embedding $sr_t^{(i)}$ of the current state. The misplaced object detector $F_d$, a binary classifier consisting of two linear layers, is utilized to fuse embeddings and detect whether the object is in a reasonable location. $det^{(i)}_t = F_d(sq^{(i)}_t, sv^{(i)}_t, sl^{(i)}_t, sr^{(i)}_t)$. $det^{(i)}_t=1$ denotes the object is misplaced in an unreasonable location and  $det^{(i)}_t=0$ denotes the object is in a reasonable location.

\subsection{Reasonable Receptacle Predictor}

The detection feature $sl^{(i)}_t$ and the commonsense embedding $sq^{(i)}_t$ are fed into an attention layer to obtain the object-commonsense attention $attl_t^{(i)}$. The room embedding $sr^{(i)}_t$ and $sq^{(i)}_t$ are also fed into an attention layer to generate the room-commonsense attention $attr_t^{(i)}$. The reasonable receptacle predictor $F_p$ including linear layers is built to fuse the visual embedding $sv^{(i)}_t$ with the attention embedding and predicts the reasonable location including receptacle $p_t$ and room type $r_t$ for the misplaced object $o_t$, then we have $(p_t, r_t) = F_p(sv^{(i)}_t, attl_t^{(i)}, attr_t^{(i)})$.

\subsection{Communication}

\begin{figure}
	\centering
	\setlength{\abovecaptionskip}{0.12cm}
	\setlength{\belowcaptionskip}{-0.55cm}
	\includegraphics[width=3.3in]{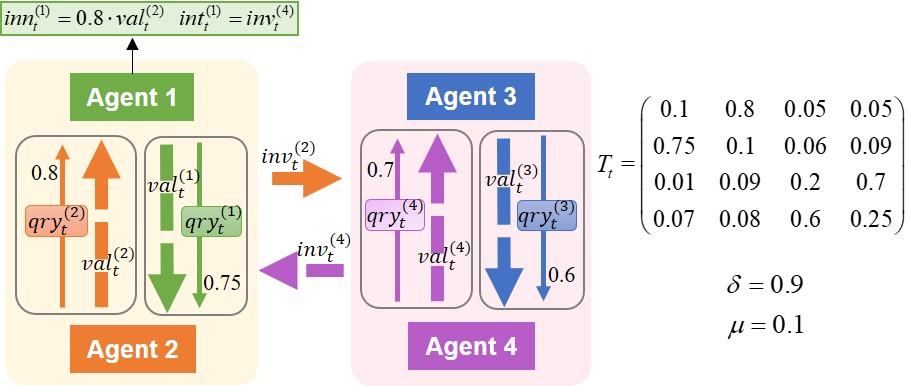}
	\caption{An example of the communication process. Based on $T_t$, at that step, \textit{Agent 1} and \textit{Agent 2} would form a communication group, \textit{Agent 3} and \textit{Agent 4} would form a group. The intra-group and inter-group communication information of $A^{(1)}$ is obtained based on $T_t$, $\delta$ and $\mu$.}
	\label{fig:group_com}
\end{figure}

Before communication, each agent generates a characteristic vector $va^{(i)}$ based on their capability and property, $va^{(i)} = (Nav^{(i)}, Mani^{(i)}, Hei^{(i)})$, and obtains a pose vector $pose_t^{(i)} = (x_t^{(i)}, y_t^{(i)}, rot_t^{(i)})$. The state feature extractor $F_s$ including linear layers and the LSTM \cite{hochreiter1997long} layer is utilized to obtain the fused state feature $sf_t^{(i)} = F_s(va^{(i)}, pose_t^{(i)}, fd_t^{(i)}, fp_t^{(i)})$, where $fd_t^{(i)}, fp_t^{(i)}$ are the extracted features from the second-from-last layers of $F_d$ and $F_p$ (the layer before the last classification layer) respectively. To solve the collaboration among heterogeneous agents, we propose the \textit{Handshake-based Group Communication (HanGrCom.)}. In this module, the agent generates the query vector $qry_t^{(i)}$, key vector $key_t^{(i)}$, value vector $val_t^{(i)}$ and inter-group information vector $inv_t^{(i)}$ based on its state feature $[qry_t^{(i)}, key_t^{(i)}, val_t^{(i)}, inv_t^{(i)}] = [\theta^q, \theta^k, \theta^v, \theta^e] (sf_t^{(i)})$, where $\theta^q$, $\theta^k$, $\theta^v$, $\theta^e$ are corresponding vector generators consisting of linear layers respectively. At the $t$-th sub-goal step, each agent sends $qry_t^{(i)}$ to others and calculates the scaled inner product attention of received query vectors and its key vector $attc_{ij} = \frac{qry_t^{(j)} key_t^{(i)}}{\sqrt{d}}$, where $d$ is the dimension of $qry_t^{(j)}$ and $key_t^{(i)}$. Then a communication matrix $T_t = \sigma([attc_{ij}]_{N\times N})$ is obtained, where $attc_{ij}$ indicates the effectiveness of the information send from $A^{(j)}$ to $A^{(i)}$, and $\sigma$ denotes row-wise softmax function. When communicating, we set two thresholds $\delta$ and $\mu$. If $attc_{ii} < \delta$, $A^{(i)}$ needs to receive messages from other agents, since its own information is not efficient. If $attc_{ij} > \mu (i \neq j)$, $A^{(i)}$ can receive messages from $A^{(j)}$. Based on $T_t$, agents can be implicitly divided into several groups. Intra-group communication exchanges the state information from agents in the same group to complete short-term sub-tasks, and $A^{(i)}$ obtain the intra-group communication information $inn_t^{(i)} = \sum_{(j \neq i,att_{ij} > \mu)} att_{ij} \cdot val_t^{(j)}$. Inter-group communication conveys higher-level information on the task allocation across different groups which is beneficial for agents to make future decisions in the long-term task completion, and $A^{(i)}$ obtains the inter-group communication information $int_t^{(i)} = inv_t^{(j)}, j=argmax(avg(att_{* j}))$, where $avg$ is the average operation and $*$ denotes the ids of agents that belong to the same group as $A^{(i)}$. $inn_t^{(i)}$ and $int_t^{(i)}$ are utilized to generate the sub-task. An example of this mechanism is shown in Fig. \ref{fig:group_com}. 

In the training process, in order to ensure the generation of communication information is differentiable, it is assumed that agents can obtain state features of other agents. At each sub-goal step, $A^{(i)}$ obtains the intra-group communication information $inn_t^{(i)} = \sum_{j=1}^{N} att_{ij} \cdot val_t^{(j)}$ and inter-group information $int_t^{(i)} = \sum_{j=1}^{N} (1 - att_{ij}) \cdot inv_t^{(j)}$. In the inference process after training, agents can only obtain their own state features and complete the communication process according to the latent divided communication groups under the threshold $\delta, \mu$ as mentioned above. We set $\mu = 0.2, \delta = 0.8$ in our experiments.

\begin{table*}[!t] 
	\setlength{\abovecaptionskip}{0.05cm}
	\small
	\caption{The Details of the Model Structure}
	
	\label{model}
	\centering
	
	\begin{tabular}{c|c}
		\Xhline{1pt}
		
		\textbf{Module} & \textbf{Layers} \\ 

		\Xhline{0.7pt}
		
		Room Classifier  & \textit{Linear} (256, 256) $\rightarrow$ \textit{Linear} (256, 4) \quad [$r_t^{(i)}$]\\
		
		\Xhline{0.7pt}
		
		Misplaced Object Detector  & \textit{Linear} (256, 256) $\rightarrow$ \textit{Linear} (256, 2) \quad [$det_t^{(i)}$]   \\ 
		
		\Xhline{0.7pt}
		
		Receptacle Predictor  & \makecell{\textit{Linear} (256, 256) $\rightarrow$ \textit{Linear} (256, 43) \quad  [$p_t^{(i)}$] \\ \qquad \qquad \qquad \quad $\searrow$  \textit{Linear} (256, 4) \qquad [$r_t^{(i)}$]} \\ 
		
		\Xhline{0.7pt}
		
		Sub-Task Predictor  & \makecell{\qquad \qquad \qquad \qquad \qquad \qquad \qquad \qquad \qquad \quad $\nearrow$ \textit{Linear} (256, 2) \quad [$Explore$ or $Place$] \\ \qquad \qquad \qquad \qquad \qquad \qquad \qquad $\nearrow$ \textit{Linear} (256, 118) \quad [$o_t^{(i)}$] \\ \textit{LSTM} (512) $\rightarrow$ \textit{Linear} (512, 256) $\rightarrow$ \textit{Linear} (256, 43) \qquad  [$p_t^{(i)}$] \\ \qquad \qquad \qquad \qquad \qquad \qquad \qquad $\searrow$ \textit{Linear} (256, 4)  \qquad \, [$r_t^{(i)}$]}  \\ 
		
		\Xhline{0.7pt}
		
		Sub-Goal Predictor  & \makecell{\qquad \qquad \qquad \qquad \qquad \qquad \quad $\nearrow$ \textit{Linear} (256, 9) \quad [$\Delta x_t^{(i)}$] \\ \qquad \qquad \qquad \qquad \qquad \qquad \quad  $\nearrow$ \textit{Linear} (256, 9) \quad [$\Delta y_t^{(i)}$] \\ \textit{LSTM} (512) $\rightarrow$ \textit{Linear} (512, 256) $\rightarrow$ \textit{Linear} (256, 5) \quad [$\Delta rot_t^{(i)}$] \\ \qquad \qquad \qquad \qquad \qquad \qquad \quad $\searrow$ \textit{Linear} (256, 4) \quad [$ope_t^{(i)}$] \\ \qquad \qquad \qquad \qquad \qquad \qquad \quad $\searrow$ \textit{Linear} (256, 2) \quad [$stop_t^{(i)}$]} \\

		\Xhline{1pt}
	\end{tabular}
\end{table*}

\subsection{Hierarchical Decision}

The hierarchical decision module generates the sub-task for each agent, predicts the sub-goal for the corresponding sub-task, and chooses the low-level actions to execute.

\begin{figure*}
	\centering
	\setlength{\abovecaptionskip}{0.1cm}
	\setlength{\belowcaptionskip}{-0.5cm}
	\includegraphics[width=6.6in]{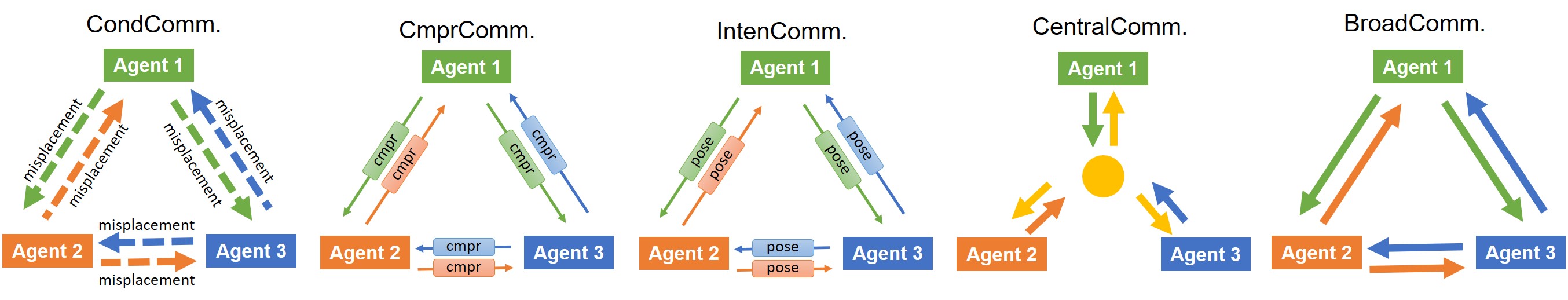}
	\caption{Different methods of communication. The thickness of the lines represents the communication amount. The thicker the line, the greater the amount. The solid line indicates the information is transmitted at each communication step, and the dotted line indicates the information is transmitted only when certain conditions are met. The rectangle on the line represents the delivered specific vectors.}
	\label{fig:comm}
\end{figure*}

\subsubsection{Sub-Task Planning}

The sub-task planning part generates the sub-task for each agent to complete. In this tidying-up task, two types of sub-tasks can be executed, \textit{Explore} and \textit{Place}. In the sub-task \textit{Explore}, the agent needs to explore the scene from its current position and search for misplaced objects. The sub-task \textit{Place} contains three parameters, which are denoted as $(Place, o_t, p_t, r_t)$, meaning to put the misplaced object $o_t$ to the receptacle $p_t$ in the room type $r_t$. A sub-task predictor $F_a$ including the LSTM layer and linear layers is utilized to predict the type and parameters of the executed sub-task with $sf_i^{(i)}$, $inn_i^{(i)}$ and $int_i^{(i)}$. We use the output of the layer before the last prediction layer to obtain the embedding $se_t^{(i)}$ of the predicted sub-task $se_t^{(i)} = F_a(sf_t^{(i)}, inn_t^{(i)}, int_t^{(i)})$.


\subsubsection{Action Decision}

Each agent encodes the $pose_t^{(i)}$ to get the pose embedding $sp_t^{(i)}$. A sub-goal predictor $F_g$ including the LSTM layer and linear layers is utilized to predict the next sub-goal $sg^{(i)}_t$ with $se_t^{(i)}$, $sm_t^{(i)}$, $sp_t^{(i)}$ and the visual embedding $sv_t^{(i)}$. $sg^{(i)}_t = (\Delta x^{(i)}_t, \Delta y^{(i)}_t, \Delta rot^{(i)}_t, ope^{(i)}_t, stop^{(i)}_t) = F_g(se_t^{(i)}, sm_t^{(i)}, sp_t^{(i)}, sv_t^{(i)})$, where $\Delta x^{(i)}_t$, $\Delta y^{(i)}_t$, $\Delta rot^{(i)}_t$ are the movement in egocentric x and y axis and the rotation angle of the agent respectively. $ope^{(i)}_t$ denotes whether performing action \textit{PickUp}, \textit{PutDown}, \textit{Drop} or \textit{NoAction} after reaching the sub-goal. $stop^{(i)}_t$ predicts whether the agent stops or not after this sub-goal step. Then, a low-level action strategy based on the shortest path algorithm is utilized to generate specific actions from $I^{(i)}$ to reach the current sub-goal.


\subsection{Model Training}

We demonstrate the details of our model structure in Table \ref{model}. $Linear(d_{in},d_{out})$ indicates the linear layer, and $d_{in}$, $d_{out}$ denote the dimensions of input and output of the layer respectively. $LSTM(d_{hid})$ indicates the LSTM layer, and $d_{hid}$ denotes the hidden size. $[*]$ represents the predicted contents. Specifically, $r_t^{(i)}$ indicates the room types which include 4 categories, $det_t^{(i)}$ indicates whether the object is misplaced or not, $p_t^{(i)}$ denotes the receptacle types which include 43 categories, $o_t^{(i)}$ denotes the object types which include 118 categories. The agent would predict the egocentric sub-goal within 1 meter in the x and y axis respectively, and the agent moves 0.25 meters each time corresponding to one grid after the discretization of the environment. The values of $\Delta x_t^{(i)}$ and $\Delta y_t^{(i)}$ are $\{-4, -3, -2, -1, 0, 1, 2, 3, 4\}$ respectively, where the negative values indicate moving in the negative direction of x and y axis, the positive values indicate moving in the positive direction of x and y axis. $rot^{(i)}_t$ can take the value of 0, 90, 180, or 270.

Since the function of each module and their training process are relatively independent, we train them separately. The misplaced object detector and reasonable receptacle predictor are trained with supervised learning, and the action decision part of the hierarchical decision module is trained with imitation learning.

We first train the room classifier in the misplaced object detection module with the true labels of the rooms where the specific object is located obtained from the houses in ProcTHOR-10K. During the training process, the parameters of the pre-trained ResNet and Mask-RCNN are frozen. After the training of the room classifier, we fix its parameters and utilize it to extract the room embedding. Then we select the pickupable objects in a number of training houses in ProcTHOR-10K and place them to reasonable or unreasonable locations in the houses to get the training samples for the misplaced object detector and the reasonable receptacle predictor. We generate the true labels of whether the corresponding objects are misplaced and extract the reasonable locations of misplaced objects based on the metadata of ProcTHOR-10K to obtain the ground truth of training samples for the misplaced object detector and the reasonable receptacle predictor respectively. We train the misplaced object detector and the reasonable receptacle predictor with the parameters of the room classifier fixed. The loss function of this process is defined as follows,
\begin{equation}\label{supervised}
	\begin{split}
		Loss_{cla} &= Loss_{type} \\ 
		Loss_{det} &= \alpha Loss_{mis} + \beta Loss_{rec1} + \lambda Loss_{room1} \\ 
	\end{split}
\end{equation}
where $Loss_{cla}$ denotes the loss function of the room classifier, and $Loss_{type}$ indicates the cross entropy loss between the predicted and the true type of the room where the specific object is located. $Loss_{det}$ denotes the loss function of the misplaced object detector and the reasonable receptacle predictor. $Loss_{mis}$ indicates the binary cross entropy of whether the object is misplaced. $Loss_{rec1}$, $Loss_{room1}$ denotes the multi-label classification loss between the predicted and true reasonable types of the receptacle and the room where the object can be placed. $\alpha$, $\beta$, $\lambda$ are the hyper-parameters. In our implementation, we set $\alpha$, $\beta$, $\lambda$ to 1.

The scene encoder consists of convolutional layers and is pre-trained to extract the map embedding of the semantic map. We design the query to predict whether a type of object exists in a sub-region of the top-down map. We generate a series of queries, which consist of a sub-area and a specific type of object, and extract the true answer of the query from the metadata of the houses. Then we pre-train the scene encoder with supervised learning, and the loss function is denoted as the entropy loss between the predicted answers and the true answers.

The hierarchical decision module predicts the next sub-task and sub-goal for each agent. The sub-task planning part predicts the type of sub-task ($Explore$ or $Place$) and the parameters of the sub-task, which contain the misplaced object type $o_t^{(i)}$, the reasonable receptacle type $p_t^{(i)}$ and the reasonable room type $r_t^{(i)}$. The action decision part predicts the next sub-goal specified by the grid of 0.25 meters. It independently predicts the number of movement grids of the sub-goal within 1 meter in the x and y axis $\Delta x^{(i)}_t$, $\Delta y^{(i)}_t$ respectively, the rotation angle $\Delta rot^{(i)}_t$, the operation action $ope^{(i)}_t$ and the probability of performing the action $Stop^{(i)}_t$. $rot^{(i)}_t$ takes the value of 0, 90, 180, or 270. $ope^{(i)}_t$ takes the value of \textit{PickUp}, \textit{PutDown}, \textit{Drop} or \textit{NoAction}. Since agents are heterogeneous and their abilities are different, the hierarchical decision modules of each agent are trained separately, and they do not share weights. We use behavior cloning to train the hierarchical decision module with the generated expert demonstrations of allocated sub-task and sub-goal actions. The loss function of training each sub-task planning part and action decision part with imitation learning is defined as follows,
\begin{equation}\label{subgoal}
	\begin{split}
		Loss_{subtask} &= Loss_{task} + \gamma_1 Loss_{obj} + \delta_1 Loss_{rec2} \\ 
		&+ \theta_1 Loss_{room2} \\
		Loss_{subgoal} &= Loss_{x_{loc}} + Loss_{y_{loc}} + \gamma_2 Loss_{rot} \\ 
		&+ \delta_2 Loss_{ope} + \theta_2 Loss_{stop} 
	\end{split}
\end{equation}
where $Loss_{task}$ indicates the binary cross entropy of the sub-task type. $Loss_{obj}$, $Loss_{rec2}$, $Loss_{room2}$ denotes the cross entropy loss between the predicted $o_t^{(i)}$, $p_t^{(i)}$, $r_t^{(i)}$ and those in demonstrations respectively. $Loss_{x_{loc}}$, $Loss_{y_{loc}}$, $Loss_{rot}$,  $Loss_{ope}$ denotes the cross entropy loss between the predicted results and those in demonstrations respectively, $Loss_{stop}$ denotes the binary cross entropy of whether the agent stops or not. $\gamma_1$, $\gamma_2$, $\delta_1$, $\delta_2$, $\theta_1$, $\theta_2$ are the hyper-parameters that can control the training process. In our implementation, we set $\gamma_1$, $\gamma_2$, $\delta_1$, $\delta_2$, $\theta_1$, $\theta_2$ to be 1.

\renewcommand{\arraystretch}{1.1}
\begin{table*}[!t] 
	\setlength{\abovecaptionskip}{0.05cm}
	\setlength{\tabcolsep}{0.55mm}
	\small
	\caption{Quantitative Results in Heterogeneous Multi-agent Tidying-up Task in \textit{Setting I}}
	
	\label{quan1}
	\centering
	
	\begin{tabular}{c|ccc|ccc|ccc|ccc|ccc|ccc}
		\Xhline{1pt}
		
		\multirow{2}{*}{Methods} & \multicolumn{3}{c|}{$Success(\uparrow)$} & \multicolumn{3}{c|}{$\%PS(\uparrow)$}& \multicolumn{3}{c|}{$\%FM(\uparrow)$}  & \multicolumn{3}{c|}{$\#PL(\downarrow)$} &\multicolumn{3}{c|}{$ACm(\downarrow)$} & \multicolumn{3}{c}{$CES(\uparrow)$} \\ 
		
		& Single  & Cross & All  & Single  & Cross & All  & Single  & Cross & All   & Single  & Cross & All & Single  & Cross & All & Single  & Cross & All \\
		
		\Xhline{0.7pt}
		SA  & 0.045  & 0.030  & 0.038  &  0.096  & 0.081  & 0.089  &  0.525  &  0.483 & 0.507  & 278.6  &  294.5  &  285.5 & - & - & - & - & - & - \\ 
		
		SA(Oracle)  & 0.090  & 0.068  & 0.080  &  0.181  & 0.163  & 0.173  & 0.703  & 0.681  & 0.693  & 184.8  & 197.1   & 190.1 & - & - & - & - & - & -\\ 
		
		\Xhline{0.7pt}
		
		Random  & 0  & 0  &  0  &  0  & 0  &  0  &  0.003  &  0  &  0.002  &  300.0  &  300.0  &  300.0 & 0 & 0 & 0 & - & - & -\\

		QMIX  & 0.032  & 0.015  &  0.025  &  0.082 &  0.069  & 0.076  & 0.472  & 0.421  & 0.450  &  293.1 & 300.0  & 296.1 &  512.0  &  512.0 &  512.0 & 0 & 0 & 0\\

		\Xhline{0.7pt}

		CondComm.  & 0.081  & 0.066  & 0.075  &  0.176  & 0.151  & 0.165  &  0.712  &  0.709 & 0.710  & 192.3  &  263.1  &  223.1 & 523.1 & 501.3 & 513.6 & 0.7 & 0.7 & 0.7 \\ 
		
		CmprComm.  & 0.067  & 0.053  & 0.061  &  0.167  & 0.131  & 0.151  &  0.695  &  0.677 & 0.687  & 188.3  &  253.3  &  216.6 & 220.0 & 220.0 & 220.0 & 1.0 & 1.0 & 1.0\\ 
		
		IntenComm.  & 0.047  & 0.032  & 0.040  &  0.136  & 0.101  & 0.121  &  0.645  &  0.609 & 0.629  & 214.6  &  267.5  &  237.6 & \textbf{20.0} & \textbf{20.0} & \textbf{20.0} & 1.0 & 1.0 & 1.0 \\ 
		
		\textbf{Ours}  & \textbf{0.121}  & \textbf{0.079}  & \textbf{0.103}  &  \textbf{0.194} & \textbf{0.160}  & \textbf{0.179}  &  \textbf{0.739}  &  \textbf{0.726} & \textbf{0.733}  & \textbf{186.1}  &  \textbf{252.6}  &  \textbf{215.0} & 363.1 & 366.5 & 364.6 & \textbf{2.1} & \textbf{1.4} & \textbf{1.8}\\ 
		
		
		\rowcolor{lightgray!30} BroadComm. & 0.130  & 0.089  & 0.112  &  0.201  & 0.184  & 0.194  & 0.742 & 0.731 & 0.737  & 183.3  &  245.9  &  210.5 & 820.0 & 820.0 & 820.0 & 1.0 & 0.7 & 0.9\\
		
		\rowcolor{lightgray!30} CentralComm.  & {0.131}  & {0.092}  & {0.114}  & {0.205}  & {0.166}  & {0.188}  &  {0.755}  & {0.739} & {0.748}  & {183.2}  &  {243.1} & {209.3} & 820.0 & 820.0 & 820.0 & 1.0 & 0.8 & 0.9 \\ 
		
		\Xhline{1pt}
	\end{tabular}
\end{table*}

\renewcommand{\arraystretch}{1.1}
\begin{table*}[!t] 
	\setlength{\abovecaptionskip}{0.05cm}
	\setlength{\tabcolsep}{0.55mm}
	\small
	\caption{Quantitative Results in Heterogeneous Multi-agent Tidying-up Task in \textit{Setting II}}
	
	\label{quan2}
	\centering
	
	\begin{tabular}{c|ccc|ccc|ccc|ccc|ccc|ccc}
		\Xhline{1pt}
		
		\multirow{2}{*}{Methods} & \multicolumn{3}{c|}{$Success(\uparrow)$} & \multicolumn{3}{c|}{$\%PS(\uparrow)$}& \multicolumn{3}{c|}{$\%FM(\uparrow)$}  & \multicolumn{3}{c|}{$\#PL(\downarrow)$} &\multicolumn{3}{c|}{$ACm(\downarrow)$} & \multicolumn{3}{c}{$CES(\uparrow)$} \\ 
		
		& Single  & Cross & All  & Single  & Cross & All  & Single  & Cross & All   & Single  & Cross & All & Single  & Cross & All & Single  & Cross & All \\
		
		\Xhline{0.7pt}
		SA  & 0.045  & 0.030  & 0.038  &  0.096  & 0.081  & 0.089  &  0.525  &  0.483 & 0.507  & 278.6  &  294.5  &  285.5 & - & - & - & - & - & - \\ 
		
		SA(Oracle)  & 0.090  & 0.068  & 0.080  &  0.181  & 0.163  & 0.173  & 0.703  & 0.681  & 0.693  & 184.8  & 197.1   & 190.1 & - & - & - & - & - & -\\ 
		
		\Xhline{0.7pt}
		
		Random  & 0  & 0  &  0  &  0  & 0  &  0  &  0.005  &  0  &  0.003  &  300.0  &  300.0  &  300.0 & 0 & 0 & 0 & - & - & -\\

		QMIX  & 0.035  & 0.021  &  0.029  &  0.091 &  0.076  & 0.084  & 0.483  & 0.425  & 0.458  &  290.5 & 296.3  & 292.9 &  512.0  &  512.0 &  512.0 & 0 & 0 & 0\\

		\Xhline{0.7pt}

		CondComm.  & 0.090  & 0.072  & 0.082  &  0.179  & 0.156  & 0.169  &  0.720  &  0.711 & 0.716  & 188.5  &  230.7  &  206.9 & 620.6 & 613.1 & 617.3 & 0.7 & 0.7 & 0.7 \\ 
		
		CmprComm.  & 0.073  & 0.059  & 0.067  &  0.173  & 0.139 & 0.158  &  0.703  &  0.683 & 0.694  & 183.1  &  226.9  &  202.1 & 330.0 & 330.0 & 330.0 & 0.8 & 0.9 & 0.8\\ 
		
		IntenComm.  & 0.048  & 0.033  & 0.041  &  0.139  & 0.112  & 0.127  &  0.661  &  0.620 & 0.643  & 210.1  &  255.3  &  229.8 & \textbf{30.0} & \textbf{30.0} & \textbf{30.0} & 1.0 & 1.0 & 1.0 \\ 
		
		\textbf{Ours}  & \textbf{0.129}  & \textbf{0.090}  & \textbf{0.112}  &  \textbf{0.203} & \textbf{0.169}  & \textbf{0.188}  &  \textbf{0.745}  &  \textbf{0.739} & \textbf{0.743}  & \textbf{166.9}  &  \textbf{221.3}  &  \textbf{190.6} & 472.5 & 478.3 & 475.0 & \textbf{1.8} & \textbf{1.3} & \textbf{1.6}\\ 
		
		
		\rowcolor{lightgray!30} BroadComm. & 0.136  & 0.094  & 0.118  &  0.209  & 0.189  & 0.200  & 0.748 & 0.741 & 0.745  & 161.1  &  213.3  &  183.8 & 1230.0 & 1230.0 & 1230.0 & 0.7 & 0.5 & 0.6\\
		
		\rowcolor{lightgray!30} CentralComm.  & {0.138}  & {0.097}  & {0.120}  & {0.211}  & {0.173}  & {0.194}  &  {0.760}  & {0.745} & {0.753}  & {160.6}  &  {213.1} & {183.4} & 1230.0 & 1230.0 & 1230.0 & 0.8 & 0.5 & 0.7 \\ 
		
		\Xhline{1pt}
	\end{tabular}
\end{table*}

\section{Experiments}\label{sec:experiment}

\subsection{Experiment Setting}

We focus on heterogeneous multi-agent collaboration, and we evaluate the proposed collaboration strategy in the tidying-up task. To effectively and efficiently train and verify our model, we select 120 scenes from the newly built dataset, 80 houses for training, 20 for validation, and 20 for testing. The number of scenes in our experiment is the same as that in other embodied tasks in AI2-THOR, which is widely used in different embodied tasks. Each house contains 10 different meta-tasks, and each meta-tasks contains 5 different initial positions for heterogeneous agents.

We consider two different heterogeneous settings with different numbers of heterogeneous agents and different ability settings. In \textit{Setting I}, there are three agents with the same visual perception ability but different action abilities and morphological characteristics. Specifically, $A^{(1)}$ only has the navigation ability with the low height, $Nav^{(1)}=1,Mani^{(1)}=0,Hei^{(1)}=0$. $A^{(2)}$ only has the navigation ability and its height is high, $Nav^{(2)}=1,Mani^{(2)}=0,Hei^{(2)}=1$. $A^{(3)}$ has both the navigation and manipulation abilities and its height is high, $Nav^{(3)}=1,Mani^{(3)}=1,Hei^{(3)}=1$. In \textit{Setting II}, there are four heterogeneous agents. The settings of $A^{(1)}$, $A^{(2)}$ and $A^{(3)}$ are the same as \textit{Setting I}, and $A^{(4)}$ has the same abilities with $A^{(3)}$. The reasons that we select the \textit{Setting I} and \textit{Setting II} are that these two settings can represent the general heterogeneous collaboration scenarios, and the experimental results with these two settings can demonstrate the generalization of our model across different numbers and heterogeneous settings of agents. In \textit{Setting I}, $A^{(3)}$ can pick up objects which are pickupable from the correct interaction locations. It is easier for $A^{(1)}$ with the lower field of view to find the objects thrown on the floor, and it can share its detected information to $A^{(3)}$ to help it pick up thrown objects. $A^{(2)}$ with the higher field of view can observe some areas that $A^{(1)}$ cannot, and its perspective can complement that of $A^{(1)}$ to assisting $A^{(3)}$ complete the tidying-up task. The visual perception of $A^{(1)}$ and $A^{(2)}$ with perspectives at different heights can complement each other during exploration, and $A^{(3)}$ can operate the misplaced objects with the assistance of $A^{(1)}$ and $A^{(2)}$ at different heights to complete the task. In \textit{Setting II}, both $A^{(3)}$ and $A^{(4)}$ can pick up misplaced objects, and they can also cooperate with each other to improve the efficiency of putting objects to the reasonable locations. These two settings can solve different situations of this task. $A^{(3)}$ ($A^{(4)}$) can pick up misplaced objects detected but cannot be picked up by $A^{(1)}$ and $A^{(2)}$. The different perspectives of $A^{(1)}, A^{(2)}$ can complement each other to easier find objects thrown on the floor. For each setting, there are a total of 4000 training tasks ($80 \times 10 \times 5$), 1000 validation tasks ($20 \times 10 \times 5$), and 1000 testing tasks. In testing tasks, there are 565 \textit{Single} tasks and 435 \textit{Cross} tasks.



\renewcommand{\arraystretch}{1.1}
\begin{table*}[!t] 
	\setlength{\abovecaptionskip}{0.05cm}
	\setlength{\belowcaptionskip}{-0.2cm}
	\setlength{\tabcolsep}{0.45mm}
	\small
	
	\caption{Ablation Experiments Results in Heterogeneous Multi-agent Tidying-up Task in \textit{Setting I}}
	
	\label{abla1}
	\centering
	
	\begin{tabular}{c|ccc|ccc|ccc|ccc|ccc|ccc}
		\Xhline{1pt}
		
		\multirow{2}{*}{Methods} & \multicolumn{3}{c|}{$ Success (\uparrow)$} & \multicolumn{3}{c|}{$\%PS (\uparrow)$}& \multicolumn{3}{c|}{$\%FM (\uparrow)$}  & \multicolumn{3}{c|}{$\#PL (\downarrow)$} &\multicolumn{3}{c|}{$ACm(\downarrow)$} & \multicolumn{3}{c}{$CES(\uparrow)$} \\ 
		
		& Single  & Cross & All  & Single  & Cross & All  & Single  & Cross & All   & Single  & Cross & All  & Single  & Cross & All  & Single  & Cross & All\\
		
		\Xhline{0.7pt}
		
		Ours w/o Know.  & 0.042  & 0.030  &  0.037  &  0.087  & 0.080  & 0.084  &  0.425  &  0.366 & 0.399  & 235.1  &  267.3  &  249.1 & 363.5 & 366.6 & 364.8 & 0 & 0 & 0 \\ 
		
		Ours w/o MisObjDec.  & 0.051  & 0.046  &  0.049  &  0.109  & 0.091  & 0.101  &  0.515  &  0.471 & 0.496  & 234.6  &  264.5  &  247.6 & 363.3 & 366.6 & 364.7 & 0.2 & 0.4 & 0.3 \\ 
		
		Ours w/o ReaRecPre.  & 0.063  & 0.058  &  0.061  &  0.120 & 0.098  & 0.110  &  0.551  &  0.509 & 0.533  & 231.1  &  260.3  &  243.8 & 363.3 & 366.9 & 364.9 & 0.5 & 0.8 & 0.6 \\ 
		
		Ours w/o Comm.  & 0.045  & 0.030  & 0.038  &  0.096  & 0.081  & 0.089  &  0.525  &  0.483 & 0.507  & 278.6  &  294.5  &  285.5 & 0 & 0 & 0 & - & - & - \\ 
		
		Ours w/o HierDec.  & 0.055  & 0.049  & 0.052  &  0.106 & 0.093  & 0.100  &  0.573  &  0.532 & 0.555  & 266.3  &  284.5  &  274.2 & 366.1 & 367.2 & 366.6 & 0.3 & 0.5 & 0.4 \\ 
		
		Ours  & {0.121}  & {0.079}  & {0.103}  &  {0.194} & {0.160}  & {0.179}  &  {0.739}  &  {0.726} & {0.733} & {186.1}  &  {252.6}  &  {215.0} & 363.1 & 366.5 & 364.6 & {2.1} & {1.4} & {1.8} \\
		
		\Xhline{1pt}
	\end{tabular}
\end{table*}

\renewcommand{\arraystretch}{1.1}
\begin{table*}[!t] 
	\setlength{\abovecaptionskip}{0.05cm}
	\setlength{\belowcaptionskip}{-0.2cm}
	\setlength{\tabcolsep}{0.45mm}
	\small
	
	\caption{Ablation Experiments Results in Heterogeneous Multi-agent Tidying-up Task in \textit{Setting II}}
	
	\label{abla2}
	\centering
	
	\begin{tabular}{c|ccc|ccc|ccc|ccc|ccc|ccc}
		\Xhline{1pt}
		
		\multirow{2}{*}{Methods} & \multicolumn{3}{c|}{$ Success (\uparrow)$} & \multicolumn{3}{c|}{$\%PS (\uparrow)$}& \multicolumn{3}{c|}{$\%FM (\uparrow)$}  & \multicolumn{3}{c|}{$\#PL (\downarrow)$} &\multicolumn{3}{c|}{$ACm(\downarrow)$} & \multicolumn{3}{c}{$CES(\uparrow)$} \\ 
		
		& Single  & Cross & All  & Single  & Cross & All  & Single  & Cross & All   & Single  & Cross & All  & Single  & Cross & All  & Single  & Cross & All\\
		
		\Xhline{0.7pt}
		
		Ours w/o Know.  & 0.049  & 0.033  &  0.042  &  0.103  & 0.096  & 0.010  &  0.452  &  0.375 & 0.419  & 227.0  &  258.5  &  240.7 & 473.3 & 479.2 & 475.9 & 0.09 & 0.06 & 0.08 \\ 
		
		Ours w/o MisObjDec.  & 0.058  & 0.052  &  0.055  &  0.118  & 0.103  & 0.111  &  0.523  &  0.480 & 0.504  & 225.5  &  256.1  &  238.8 & 473.1 & 479.0 & 475.7 & 0.3 & 0.5 & 0.4 \\ 
		
		Ours w/o ReaRecPre.  & 0.075  & 0.069  &  0.072  &  0.128 & 0.105  & 0.118  &  0.560  &  0.515 & 0.540  & 220.6  &  251.9  &  234.2 & 473.3 & 478.9 & 475.8 & 0.6 & 0.8 & 0.7 \\ 
		
		Ours w/o Comm.  & 0.049  & 0.033  & 0.042  &  0.105  & 0.091  & 0.099  &  0.535  &  0.496 & 0.518  & 273.1  &  280.6  &  276.4 & 0 & 0 & 0 & - & - & - \\ 
		
		Ours w/o HierDec.  & 0.063  & 0.055  & 0.060  &  0.111 & 0.99  & 0.106  &  0.581  &  0.537 & 0.562  & 260.1  &  273.6  &  266.2 & 475.3 & 478.8 & 477.3 & 0.4 & 0.5 & 0.4 \\ 
		
		Ours  & {0.129}  & {0.090}  & {0.112}  &  {0.203} & {0.169}  & {0.188}  &  {0.745}  &  {0.739} & {0.743}  & {166.9}  &  {221.3}  & {190.6} & 472.5 & 478.3 & 475.0 & {1.8} & {1.3} & {1.6} \\
		
		\Xhline{1pt}
	\end{tabular}
\end{table*}

\subsection{Evaluation Metrics}

We propose six metrics to evaluate the performance of the model, and $N_{epi}$ denotes the number of tasks: 

\textit{1) Success (Suc)}: $Suc=\frac{1}{N_{epi}}\sum_{i=1}^{N_{epi}}R_i$, where $R_i=1$ if all misplaced objects are re-placed in reasonable locations, otherwise $R_i=0$. It denotes the strict success rate.

\textit{2) \%PatialSuccess (\%PS)}: $\%PS=\frac{1}{N_{epi}}\sum_{i=1}^{N_{epi}}\frac{K^{suc}_i}{K_i}$, where ${K_i}$ denotes the number of misplaced objects in the $i$-th task, and $K^{suc}_i$ is the number of misplaced objects that are re-placed to reasonable locations by agents in this task. It can reflect the proportion of misplaced objects that can be successfully tidied in each task.

\textit{3) \%FindMisObj (\%FM)}: $\%FM=\frac{1}{N_{epi}}\sum_{i=1}^{N_{epi}}\frac{K^{det}_i}{K_i}$, where $K^{det}_i$ is the number of misplaced objects that are detected and picked up by agents in this task. It can reflect the effect of misplaced objects detector to some extent.

\textit{4) \#PathLength (\#PL)}: $\#PL=\frac{1}{N_{epi}}\sum_{i=1}^{N_{epi}}Len_i$, where $Len_i$ is the number of steps in the multi-agent trajectory when synchronously completing the $i$-th task, that is, the maximum number of steps of three agents. It evaluates the efficiency of completing the task. The fewer steps, the higher efficiency of the multi-agent collaboration.

\textit{5) AvgCom (ACm)}: $ACm=\frac{1}{N_{epi}}\sum_{i=1}^{N_{epi}} \frac{Total_i}{Len_i \cdot N}$, where $Total_i$ is the total dimensions of communication messages among agents. It evaluates the average communication amounts per agent per step.

\textit{6) CommunicationEfficiencyScore (CES)}: $CES=max(0, \frac{10000 \cdot (Suc - SucSA)}{ACm})$, where $SucSA$ is the success rate of the single-agent model. It evaluates the proportion of efficiency improvement of different communication methods. The higher $CES$, the higher efficiency of the communication mechanism.

\subsection{Quantitative Analysis}

We evaluate the proposed model with different action decision strategies and communication methods in testing tasks. For different action strategies, we compare the results with \textit{SA}, \textit{SA(Oracle)}, \textit{Random}, and \textit{QMIX} model. \textit{SA} only utilizes one agent with both navigation and manipulation abilities to complete the task, and the other parts are the same as the proposed model. In \textit{SA(Oracle)}, the objects' locations are known to the agent, and the other parts are the same as \textit{SA}. \textit{Random} model randomly generates actions for each agent from their action spaces. \textit{QMIX} is a MARL method based on \cite{rashid2020monotonic}, in which agents obtain all state features from others, fuse with their own state features, and directly predict the next low-level actions based on fused state features. For different communication methods, we consider several methods that save bandwidth. In \textit{CondComm.}, only when an agent detects the misplaced object in the current view, it would broadcast its state and map to others. \textit{CmprComm.} utilizes an encoder-decoder structure to compress the communication messages, and decode received messages from other agents, similar to that in \cite{liu2022multi}. In \textit{IntenComm.}, agents exchange state and sub-task information, infer other agent's next sub-goal, and then decide their next sub-task, which is referred to as the idea in \cite{wang2021tom2c}. We also pay attention to the central and broadcast mechanisms, which require large communication amounts. \textit{CentralComm.} adopts a central node to obtain the messages from all agents and generates the sub-task for each agent. In \textit{BroadComm.}, each agent broadcasts its messages to all agents and determines the next actions with its state and received messages from others. All these communication baselines are demonstrated in Fig. \ref{fig:comm}.

\begin{figure*} [t!]
	\centering
	\subfloat[The success sample in \textit{Setting I} with three agents.]{\label{a}
		\includegraphics[width=7.1in]{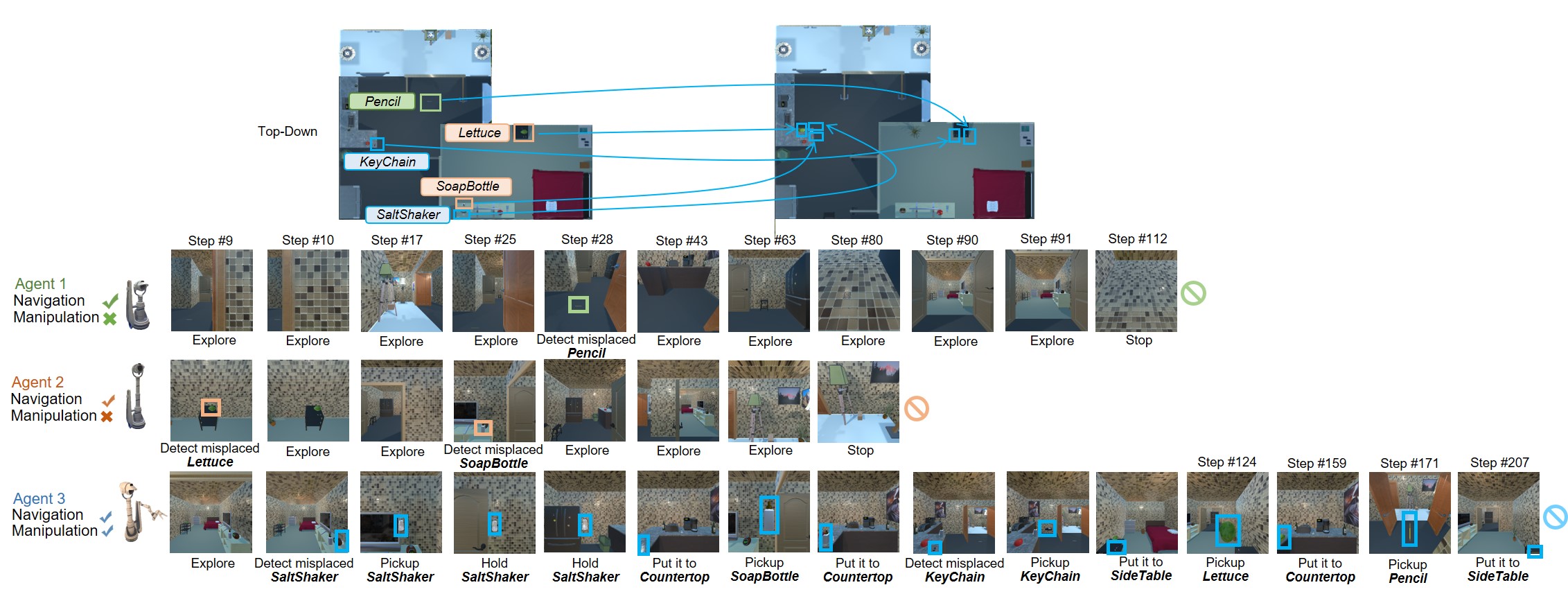}}
	\\
	\hspace{0.2in}
	\subfloat[The success sample in \textit{Setting II} with four agents.]{\label{b}
		\includegraphics[width=7.1in]{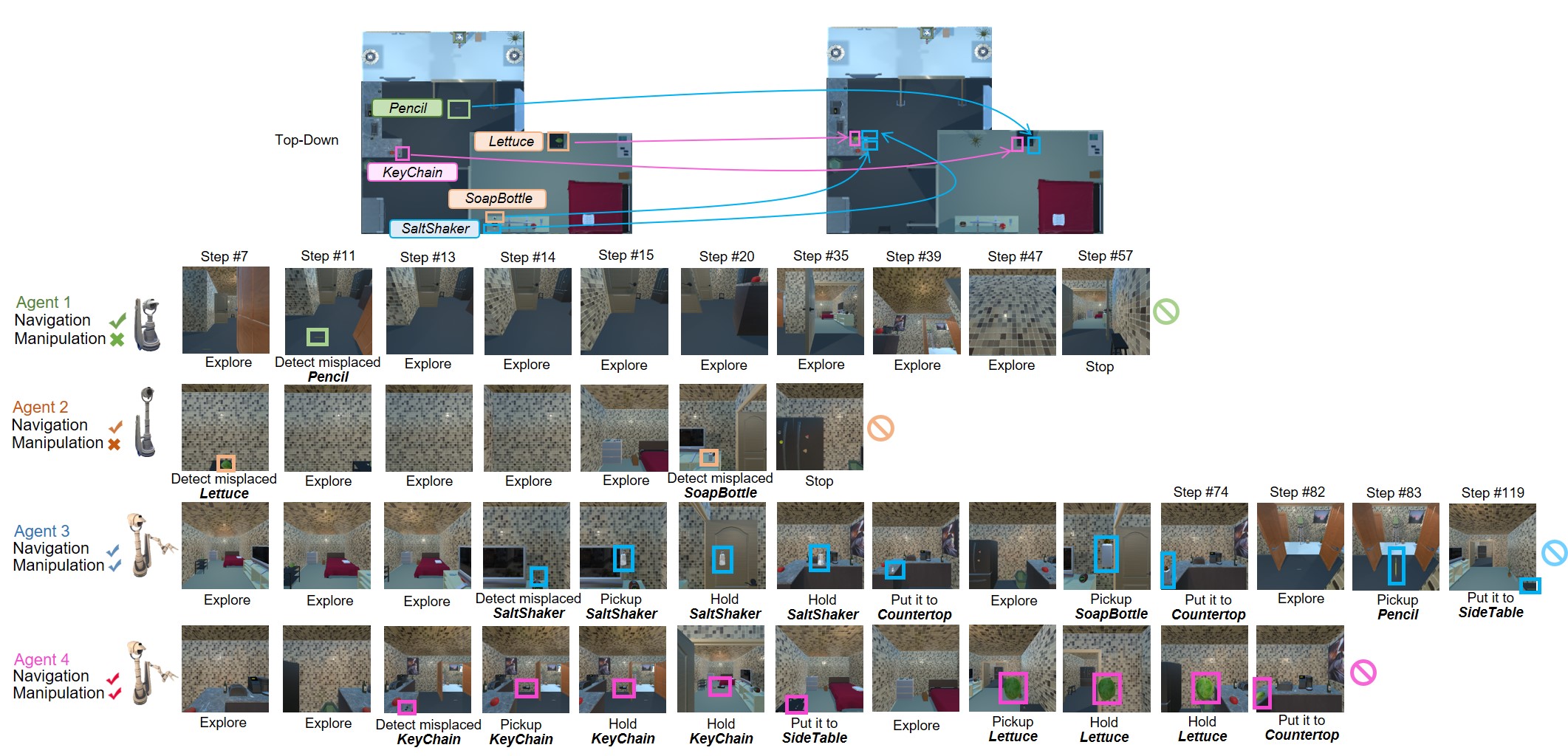} }
	\caption{The successful samples in the same house in \textit{Setting I} and \textit{Setting II}. In the top-down view, the color of bounding boxes in the left indicates which agent detects the specific misplaced objects, and the color of arrows indicates which agent puts the misplaced objects to reasonable locations.}
	\label{fig:succ} 
\end{figure*}

\begin{figure}
	\centering
	\includegraphics[width=3.2in]{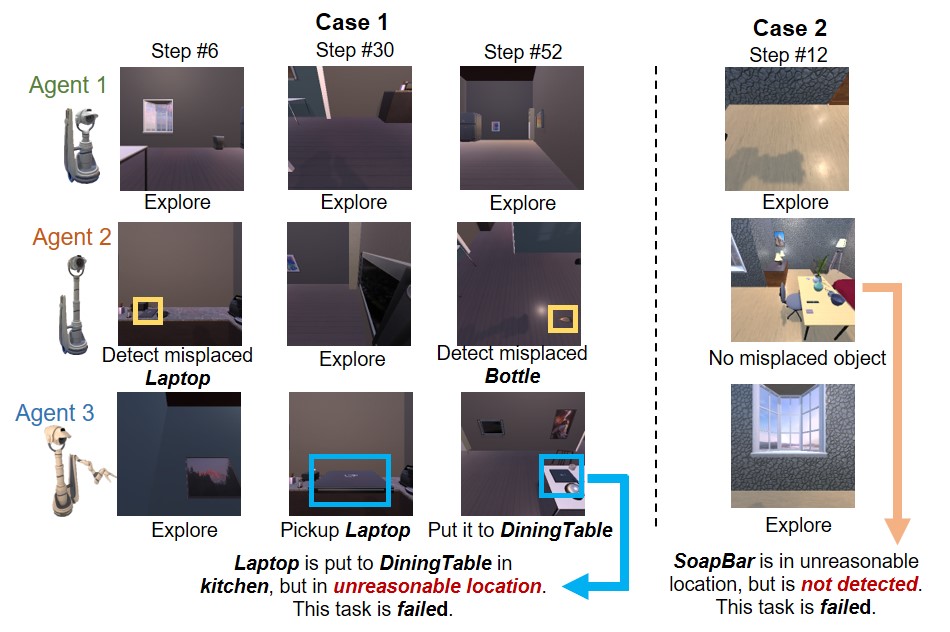}
	\caption{The failed cases in \textit{Setting I} of the task.In \textit{Case 1}, \textit{Agent 3} place \textit{Laptop} to the \textit{Kitchen}, an unreasonable location. In \textit{Case 2}, \textit{Agent 2} fail to detect \textit{SoapBar} in the \textit{DiningTable} in \textit{Bedroom}.}
	\label{fig:failcase}
\end{figure}

We show the quantitative results of \textit{Single} and \textit{Cross} tasks in \textit{Setting I}, \textit{Setting II} in Table \ref{quan1} and Table \ref{quan2} respectively. The best performance except for central and broadcast methods are bolded. The maximum number of steps per task is 300. The results show that the heterogeneous multi-agent tidying-up task is challenging. $\%FM$ is relatively high, but the success rate of the whole task is relatively low, which indicates that moving the detected misplaced objects to the reasonable location is challenging, since agents may not reason the proper receptacle or not generate the valid manipulation actions to re-place the misplaced object to predicted locations. The results in \textit{Setting I} and \textit{Setting II} indicate that our model can generalize to different numbers and heterogeneous settings of agents. Compared with multi-agent models, the success rate of task completion of \textit{SA} and \textit{SA(Oracle)} is lower, which demonstrates that collaboration and communication among heterogeneous multi-agents can help agents improve the accuracy and efficiency of tasks. Although agents in \textit{QMIX} can obtain state features of other agents, its performance is relatively low, indicating that this MARL method is not effective enough to deal with visual information, and it is difficult to directly predict low-level actions.

As for different communication methods, the performances of \textit{Ours}, \textit{CondComm.}, \textit{CmprComm.} and \textit{IntenComm.} in task completion ($Suc$,$\%PS$,$\%FM$,$\#PL$) are lower than \textit{CentralComm.} and \textit{BroadComm.}, because agents share partial information with others for decision-making. Since \textit{CentralComm.} utilizes a central node to process information, its task completion performance can be regarded as the upper bound of multi-agent models. Our model performs better than other communication methods that share partial messages with agents in task completion. In our model, the agent with manipulation ability can obtain the state and map information from other agents with only the navigation ability, which is beneficial for the agent to put the object to a reasonable location, and its performance is better than the other three models. In \textit{CondComm.}, agents exchange their state only when a misplaced object is detected, and the incomplete shared map information makes its performance lower than our model. The performance of \textit{CmprComm.} indicates that decoding the compressed communication message would lose some effective information though the communication bandwidth can be reduced. The performance of \textit{IntenComm.} is relatively low due to the improper reasoning results of other agents' decisions and incomplete exchanging of information. Furthermore, in \textit{QMIX}, each agent receives the fused 512-dimensional state features from other agents, so $ACm$ of \textit{QMIX} is always 512. The size of local map embedding $sm_t^{(i)}$ is $20 \times 20$. The transmitted state information contains the 3-dimensional $va^{(i)}$, 3-dimensional pose $pose_t^{(i)}$ and 4-dimensional sub-task information $(Place/Explore, o_t, p_t, r_t)$, so the number of total dimensions of the state information from one agent is 10. Since \textit{IntenComm.} only transmits the state information from other agents to each agent, so \textit{ACm} in \textit{Setting I} is 20 and \textit{ACm} in \textit{Setting II} is 30. In \textit{CmprComm.}, each agent compresses the dimensions of the map embedding to be 100, and transmits the compressed map embedding as well as the state information to other agents, so \textit{ACm} in \textit{Setting I} is $2 \times (100 + 10) = 220$, \textit{ACm} in \textit{Setting II} is $3 \times (100 + 10) = 330$. In \textit{BroadComm.} and \textit{CentralComm.}, each agent transmits their state information and map embeddings to all other agents, so \textit{ACm} in \textit{Setting I} is $2 \times (400 + 10) = 820$ and \textit{ACm} in \textit{Setting II} is $3 \times (400 + 10) = 1230$. $ACm$ of \textit{CentralComm.} and \textit{BroadComm.} are the largest, and with the increased number of heterogeneous agents, their $ACm$ have increased greatly, but the declines in $SCE$ are relatively larger, which indicates that they are not cost-effective enough. Although \textit{IntenComm.} uses the smallest $ACm$, its $CES$ is not high, since its $Suc$ is not good. Our method can save communication amounts, achieve good performance of task completion, and obtain the highest communication efficiency $CES$, demonstrating the effectiveness of the proposed communication mechanism.

\subsection{Qualitative Analysis}

We demonstrate successful samples with three agents in \textit{Setting I} and four agents in \textit{Setting II} respectively in Fig. \ref{fig:succ}. The tidying-up tasks in \ref{a} and \ref{b} are the same. Agents complete the same task more quickly in \textit{Setting II}, since \textit{Agent 3} and \textit{Agent 4} with both the navigation and manipulation abilities can collaborate with each other to pick up misplaced objects and put them to reasonable locations, which demonstrates that the proposed model can make effective use of different abilities of heterogeneous to improve efficiency. We also show two failed cases in Fig. \ref{fig:failcase}, where the agent places the misplaced object in an unreasonable location or agents fail to detect all the misplaced objects.

\subsection{Ablation Experiments}
To evaluate the role and effectiveness of different modules of our model, we conduct ablation experiments by removing the specific part of the model and comparing the performance. \textit{Ours w/o Know.} removes the commonsense prior knowledge, which judge the misplaced objects and infer the reasonable locations directly from visual features. \textit{Ours w/o MisObjDec.} removes the misplaced object detector, which extracts visual features and relationship features from the current observation as the subsequent input of the model. \textit{Ours w/o ReaRecPre.} removes the reasonable receptacle predictor and directly extracts the receptacle features with the linear layer to input to the subsequent model to generate next actions. We remove the communication module to obtain \textit{Ours w/o Comm.}, in which agents do not communicate with others and each agent performs actions independently. \textit{Ours w/o HierDec.} removes the hierarchical decision, and agents do not predict the executing sub-tasks and sub-goals, but directly generate low-level actions with their state features. The results of the ablation experiments for \textit{Single} and \textit{Cross} tasks in \textit{Setting I}, \textit{Setting II} in Table \ref{abla1} and Table \ref{abla2} respectively.


The performance of \textit{Ours w/o Know.} drops obviously, indicating that the commonsense knowledge is quite important for scene reasoning in this task. Since the misplaced object detector and the reasonable receptacle predictor can provide a clear target object to be picked up and a clear receptacle to put the object for subsequent sub-tasks generating, the performance of \textit{Ours w/o MisObjDec.} and \textit{Ours w/o ReaRecPre.} would decrease. In \textit{Ours w/o Comm.}, each agent has no communication with others, since in \textit{Setting I} only one agent has the manipulation capability, the performance is the same as \textit{SA} in \textit{Setting I}. \textit{Ours w/o HierDec.} directly generates the low-level actions with the detected misplaced object, the reasoned receptacle, and the state features. Because it is difficult to directly learn the relations between state features and executed low-level actions, this model would generate some invalid actions for the executing task, which leads to worse performance. Since the misplaced object detector and reasonable receptacle predictor are retained in \textit{Ours w/o HierDec.}, its $\%FM$ is still higher than that of \textit{Ours w/o MisObjDec.} and \textit{Ours w/o ReaRecPre.}. The results demonstrate that each module of the proposed model is effective for heterogeneous multi-agent task completion.

\section{Conclusion}\label{sec:conclusion}

In this paper, we propose the heterogeneous multi-agent collaborative framework based on the handshake-based group communication strategy and hierarchical decision model. To evaluate the effectiveness of the framework, we propose the heterogeneous multi-agent tidying-up task and generate a benchmark dataset for this task in houses with multiple rooms. The results demonstrate the effectiveness of each module of our model. In the future, we will continue to study the communication mechanism in more complex tasks and larger multi-agent systems with both homogeneous and heterogeneous agents.

\ifCLASSOPTIONcompsoc
  \section*{Acknowledgments}
\else
  \section*{Acknowledgment}
\fi

This work was supported in part by the National Natural Science Foundation of China under Grants 62025304.

\ifCLASSOPTIONcaptionsoff
  \newpage
\fi



\bibliographystyle{IEEEtran}
\bibliography{main}

\end{document}